\documentclass[twoside,twocolumn,a4paper,11pt,fleqn]{article}
\usepackage{colacl}
\usepackage{latexsym}
\newcommand{\brover}[1]{%
\makebox[0pt][l]{$\overbrace{\rule{0pt}{1.5ex}\hspace{#1}}$}%
}%
\newcommand{\brunder}[1]{%
\makebox[0pt][l]{\raisebox{-.5ex}{$\underbrace{\hspace{#1}}$}}%
}%
\usepackage{hhline}
\begin{document}
\title{\textbf{ABL: Alignment-Based Learning}}
\author{Menno~van~Zaanen\\
School of Computer Studies\\
University of Leeds\\
LS2 9JT  Leeds\\
UK\\
\texttt{menno@scs.leeds.ac.uk}}
\date{}
\maketitle

\begin{abstract}
This paper introduces a new type of grammar learning algorithm, inspired by
string edit distance \cite{bib:tstscp}. The algorithm takes a corpus of
flat sentences as input and returns a corpus of labelled, bracketed
sentences. The method works on pairs of unstructured sentences that have one or
more words in common. When two sentences are divided into parts that are the
same in both sentences and parts that are different, this information is used
to find parts that are interchangeable. These parts are taken as possible
constituents of the same type. After this alignment learning step, the
selection learning step selects the most probable constituents from all
possible constituents.

This method was used to bootstrap structure on the ATIS corpus
\cite{bib:balacoetpt} and on the OVIS\footnote{Openbaar Vervoer Informatie
Systeem (OVIS) stands for Public Transport Information System.} corpus
\cite{bib:admfsi}. While the results are encouraging (we obtained up to 89.25~\%
non-crossing brackets precision), this paper will point out some of the
shortcomings of our approach and will suggest possible solutions.  
\end{abstract}

\section{Introduction}

Unsupervised learning of syntactic structure is one of the hardest problems in
NLP. Although people are adept at learning grammatical structure, it is
difficult to model this process and therefore it is hard to make a computer
learn structure.

We do not claim that the algorithm described here models the human process of
language learning. Instead, the algorithm should, given unstructured sentences,
find the best structure. This means that the algorithm should assign structure
to sentences which is similar to the structure people would give to sentences,
but not necessarily in the same time or space restrictions.

The algorithm consists of two phases. The first phase is a constituent
generator, which generates a motivated set of possible constituents by aligning
sentences. The second phase restricts this set by selecting the best
constituents from the set.

The rest of this paper is organized as follows. Firstly, we will start by
describing previous work in machine learning of language structure and then we
will give a description of the ABL algorithm. Next, some results of applying
the ABL algorithm to different corpora will be given, followed by a discussion
of the algorithm and future research.

\section{Previous Work}

Learning methods can be grouped into supervised and unsupervised methods.
Supervised methods are initialised with structured input (i.e. structured
sentences for grammar learning methods), while unsupervised methods learn by
using unstructured data only.

In practice, supervised methods outperform unsupervised methods, since
they can adapt their output based on the structured examples in the
initialisation phase whereas unsupervised methods cannot. However, it is
worthwhile to investigate unsupervised grammar learning methods, since ``the
costs of annotation are prohibitively time and expertise intensive, and the
resulting corpora may be too susceptible to restriction to a particular domain,
application, or genre''. \cite{bib:ulinlp}

There have been several approaches to the \emph{unsupervised} learning of
syntactic structures. We will give a short overview here.

Memory based learning (MBL) keeps track of possible contexts and assigns
word types based on that information \cite{bib:mblaap}.
\newcite{bib:diapcfasc} present a method that bootstraps syntactic categories
using distributional information and \newcite{bib:pnlumis} describe a method
that finds constituent boundaries using mutual information values of the part of
speech n-grams within a sentence.

Algorithms that use the minimum description length (MDL) principle build
grammars that describe the input sentences using the minimal number of bits.
This idea stems from information theory. Examples of these systems can be
found in \cite{bib:giatmdlp} and \cite{bib:ula}.

The system by \newcite{bib:ladcag} performs a heuristic search
while creating and merging symbols directed by an evaluation function.
\newcite{bib:bgiflm} presents a Bayesian grammar induction method, which is
followed by a post-pass using the inside-outside algorithm
\cite{bib:tgfsr,bib:teoscfgutioa}.

Most work described here cannot learn complex structures such as recursion,
while other systems only use limited context to find constituents. However,
the two phases in ABL are closely related to some previous work. The alignment
learning phase is effectively a compression technique comparable to MDL or
Bayesian grammar induction methods. ABL remembers all possible constituents,
building a search space. The selection learning phase searches this space,
directed by a probabilistic evaluation function.

\section{Algorithm}

We will describe an algorithm that learns structure using a corpus of plain
(unstructured) sentences. It does not need a structured training set to
initialize, all structural information is gathered from the unstructured
sentences.

The output of the algorithm is a labelled, bracketed version of the input
corpus. Although the algorithm does not generate a (context-free) grammar, it
is trivial to deduce one from the structured corpus.

The algorithm builds on Harris's idea \shortcite{bib:misl} that states that
\emph{constituents of the same type can be replaced by each other}. Consider the
sentences as shown in figure~\ref{fig:sentences}.\footnote{All sentences in the
examples can be found in the ATIS corpus.} The constituents \textit{a family
fare} and \textit{the payload of an African Swallow} both have the same
syntactic type (they are both NPs), so they can be replaced by each other. This
means that when the constituent in the first sentence is replaced by
the constituent in the second sentence, the result is a valid sentence in
the language; it is the second sentence.

\begin{figure}
\begin{tabular}{l}
\textit{\underline{What is} a family fare}\\
\textit{\underline{What is} the payload of an African Swallow}\\[.05cm]
\hline
\textit{What is (a family fare)$_X$}\\
\textit{What is (the payload of an African Swallow)$_X$}
\end{tabular}
\caption{Example bootstrapping structure}
\label{fig:sentences}
\end{figure}

The main goal of the algorithm is to establish that \textit{a family fare} and
\textit{the payload of an African Swallow} are constituents and have the same
type. This is done by reversing Harris's idea: \emph{if (a group of) words can
be replaced by each other, they are constituents and have the same type}. So
the algorithm now has to find groups of words that can be replaced by each
other and after replacement still generate valid sentences. 

The algorithm consists of two steps:
\begin{enumerate}
\item Alignment Learning
\item Selection Learning
\end{enumerate}

\subsection{Alignment Learning}

The model learns by comparing all sentences in the input corpus to each other
in pairs. An overview of the algorithm can be found in
figure~\ref{fig:algorithm}.

\begin{figure}
\begin{tabbing}
For\=For\=For\=\kill
For each sentence $s_1$ in the corpus:\\
\>For every other sentence $s_2$ in the corpus:\\
\>\>Align $s_1$ to $s_2$\\
\>\>Find the identical and distinct parts\\
\>\>\>between $s_1$ and $s_2$\\
\>\>Assign non-terminals to the constituents\\
\>\>\>(i.e. distinct parts of $s_1$ and $s_2$)
\end{tabbing}
\caption{Alignment learning algorithm}
\label{fig:algorithm}
\end{figure}

Aligning sentences results in ``linking'' identical words in the sentences.
Adjacent linked words are then grouped. This process reveals the groups of
identical words, but it also uncovers the groups of distinct words in the
sentences. In figure~\ref{fig:sentences} \textit{What is} is the identical part
of the sentences and \textit{a family fare} and \textit{the payload of an
African Swallow} are the distinct parts. The distinct parts are
interchangeable, so they are determined to be constituents of the same type.

We will now explain the steps in the alignment learning phase in more detail.

\subsubsection{Edit Distance}

\begin{figure}
\begin{tabular}{l}
\textit{\underline{from} ()$_1$ \underline{San Francisco} (to Dallas)$_2$}\\
\textit{\underline{from} (Dallas to)$_1$ \underline{San Francisco}
()$_2$}\\[.5cm]
\textit{\underline{from} (San Francisco to)$_1$ \underline{Dallas} ()$_2$}\\
\textit{\underline{from} ()$_1$ \underline{Dallas} (to San
Francisco)$_2$}\\[.5cm]
\textit{\underline{from} (San Francisco)$_1$ \underline{to} (Dallas)$_2$}\\
\textit{\underline{from} (Dallas)$_1$ \underline{to} (San Francisco)$_2$}\\
\end{tabular}
\caption{Ambiguous alignments}
\label{fig:link}
\end{figure}

To find the identical word groups in the sentences, we use the edit distance
algorithm by Wagner and Fischer \shortcite{bib:tstscp}, which finds the minimum
number of edit operations (insertion, deletion and substitution) to change one
sentence into the other. Identical words in the sentences can be found at
places where no edit operation was applied.

The instantiation of the algorithm that finds the longest common subsequence in
two sentences sometimes ``links'' words that are too far apart. In
figure~\ref{fig:link} when, besides the occurrences of \textit{from}, the
occurrences of \textit{San Francisco} or \textit{Dallas} are linked, this
results in unintended constituents. We would rather have the model linking
\textit{to}, resulting in a structure with the noun phrases grouped
with the same type correctly.

Linking \textit{San Francisco} or \textit{Dallas} results in constituents that
vary widely in size. This stems from the large distance between the linked
words in the first sentence and in the second sentence. This type of
alignment can be ruled out by biasing the cost function using distances
between words.

\subsubsection{Grouping}

An edit distance algorithm links identical words in two sentences. When adjacent
words are linked in both sentences, they can be grouped. A group like this is a
part of a sentence that can also be found in the other sentence. (In
figure~\ref{fig:sentences}, \textit{What is} is a group like this.)

The rest of the sentences can also be grouped. The words in these groups are
words that are distinct in the two sentences. When all of these groups from
sentence one would be replaced by the respective groups of sentence two,
sentence two is generated. (\textit{a family fare} and \textit{the
payload of an African Swallow} are of this type of group in
figure~\ref{fig:sentences}.) Each pair of these distinct groups consists of
possible constituents of the same type.\footnote{Since the algorithm does not
know any (linguistic) names for the types, the algorithm chooses natural
numbers to denote different types.}

As can be seen in figure~\ref{fig:link}, it is possible that empty groups can
be learned.

\subsubsection{Existing Constituents}
\label{s:ec}

At some point it may be possible that the model learns a constituent that was
already stored. This may happen when a new sentence is compared to a sentence
in the partially structured corpus. In this case, no new type is introduced,
but the constituent in the new sentence gets the same type of the constituent
in the sentence in the partially structured corpus.

It may even be the case that a partially structured sentence is compared to
another partially structured sentence. This occurs when a sentence that contains
some structure, which was learned by comparing to a sentence in the partially
structured corpus, is compared to another (partially structured) sentence. When
the comparison of these two sentences yields a constituent that was already
present in both sentences, the types of these constituents are merged. All
constituents of these types are updated, so they have the same type.

By merging types of constituents we make the assumption that constituents in a
certain context can only have one type. In section~\ref{s:wst} we discuss the
implications of this assumption and propose an alternative approach.

\subsection{Selection Learning}

\begin{figure}
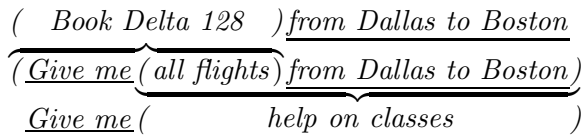

\begin{tabular}{c@{\hspace{0pt}}c@{\hspace{1pt}}c@{\hspace{0pt}}c@{\hspace{0pt}}c@{\hspace{1pt}}c@{\hspace{0pt}}c}
\textit{(} & \multicolumn{3}{c}{\textit{Book Delta 128}} & \textit{)} &
\underline{\textit{from Dallas to Boston}}\\
\brover{3.65cm}\textit{(} & \underline{\textit{Give me}} &
\brunder{5.9cm}\textit{(} & \textit{all flights} & ) &
\underline{\textit{from Dallas to Boston}} & \textit{)}\\
& \underline{\textit{Give me}} & \textit{(} &
\multicolumn{3}{c}{\textit{help on classes}} & \textit{)}\\
\end{tabular}
\caption{Overlapping constituents}
\label{fig:overlap}
\end{figure}

The first step in the algorithm may at some point generate constituents that
overlap with other constituents. In figure~\ref{fig:overlap} \textit{Give me all
flights from Dallas to Boston} receives two overlapping structures. One
constituent is learned by comparing against \textit{Book Delta 128 from Dallas
to Boston} and the other (overlapping) constituent is found by aligning with
\textit{Give me help on classes}.

The solution to this problem has to do with selecting the correct constituents
(or at least the better constituents) out of the possible constituents.
Selecting constituents can be done in several different ways.

\setlength{\leftmargini}{1em}
\begin{description}
\item [ABL:incr] Assume that the first constituent learned is the
correct one. This means that when a new constituent overlaps with older
constituents, it can be ignored (i.e. they are not stored in the corpus).
\item [ABL:leaf] The model computes the probability of a constituent counting
the number of times the particular words of the constituent have occurred in
the learned text as a constituent, normalized by the total number of
constituents.
\[P_{leaf}(c)=\frac{|c'\in C:yield(c')=yield(c)|}{|C|}\] where $C$ is the
entire set of constituents.
\item [ABL:branch] In addition to the words of the sentence delimited by the
constituent, the model computes the probability based on the part of the
sentence delimited by the words of the constituent \emph{and} its non-terminal
(i.e. a normalised probability of ABL:leaf).
\end{description}
\vspace{-.5cm}
\begin{eqnarray*}
\lefteqn{P_{branch}(c|root(c)=r)=}\\
& &\frac{|c'\in C:yield(c')=yield(c)\wedge root(c')=r|}{|c''\in C:root(c'')=r|}
\end{eqnarray*}

The first method is non-probabilistic and may be applied every time a
constituent is found that overlaps with a known constituent (i.e. while
learning).

The two other methods are probabilistic. The model computes the probability
of the constituents and then uses that probability to select constituents with
the highest probability. These methods are applied after the alignment
learning phase, since more specific information (in the form of better counts)
can be found at that time.

In section~\ref{s:results} we will evaluate all three methods on the ATIS and
OVIS corpus.

\subsubsection{Viterbi}

Since more than just two constituents can overlap, all possible
combinations of overlapping constituents should be considered when
computing the best combination of constituents, which is the product of the
probabilities of the separate constituents as in SCFGs (cf. \cite{bib:profl}).
A Viterbi style algorithm optimization \shortcite{bib:ebfccaaaoda} is used to
efficiently select the best combination of constituents.

When computing the probability of a combination of constituents, multiplying the
separate probabilities of the constituents biases towards a low number of
constituents. Therefore, we compute the probability of a set of constituents 
using a normalized version, the geometric mean\footnote{The geometric mean of
a set of constituents $c_1, \ldots, c_n$ is \(P(c_1\wedge\ldots\wedge
c_n)=\sqrt[n]{\prod_{i=1}^n P(c_i)}\)}, rather than its product.
\cite{bib:nfomfbfpcp}

\section{Results}
\label{s:results}

\begin{table*}
\begin{center}
\begin{small}
\begin{sc}
\begin{tabular}
{|l@{\hspace{3pt}}|
r@{\hspace{3pt}}r@{\hspace{3pt}}|
r@{\hspace{3pt}}r@{\hspace{3pt}}|
r@{\hspace{3pt}}r@{\hspace{3pt}}||
r@{\hspace{3pt}}r@{\hspace{3pt}}|
r@{\hspace{3pt}}r@{\hspace{3pt}}|
r@{\hspace{3pt}}r@{\hspace{3pt}}|
}\cline{2-13}
\multicolumn{1}{c|}{}
& \multicolumn{6}{c||}{ATIS}
& \multicolumn{6}{c|}{OVIS}\\\cline{2-13}
\multicolumn{1}{c|}{}
& \multicolumn{2}{c|}{NCBP}
& \multicolumn{2}{c|}{NCBR}
& \multicolumn{2}{c||}{ZCS}
& \multicolumn{2}{c|}{NCBP}
& \multicolumn{2}{c|}{NCBR}
& \multicolumn{2}{c|}{ZCS}\\\hline
left       & 32.60 &        & 76.82 &        &  1.12 &
           & 51.23 &        & 73.17 &        & 25.22 &        \\
right      & 82.70 &        & 92.91 &        & 38.83 &
           & 75.85 &        & 86.66 &        & 48.08 &        \\
ABL:incr   & 83.24 & (1.17) & 87.21 & (0.67) & 18.56 & (2.32)
           & 88.71 & (0.79) & 84.36 & (1.10) & 45.11 & (3.22) \\
ABL:leaf   & 81.42 & (0.11) & 86.27 & (0.06) & 21.63 & (0.50)
           & 85.32 & (0.02) & 79.96 & (0.03) & 30.87 & (0.09) \\
ABL:branch & 85.31 & (0.01) & 89.31 & (0.01) & 29.75 & (0.00)
           & 89.25 & (0.00) & 85.04 & (0.00) & 42.20 & (0.01) \\\hline
\end{tabular}
\end{sc}
\end{small}
\end{center}
\vspace{-.3cm}
\caption{Results of the ATIS and OVIS corpora}
\label{fig:results}
\vspace{-.2cm}
\end{table*}

The three different ABL algorithms and two baseline systems have been tested on
the ATIS and OVIS corpora.

The ATIS corpus from the Penn Treebank consists of 716 sentences containing
11,777 constituents. The larger OVIS corpus is a Dutch corpus containing
sentences on travel information. It consists of exactly 10,000 sentences. We
have removed all sentences containing only one word, resulting in a corpus of
6,797 sentences and 48,562 constituents.

The sentences of the corpora are stripped of their structures. These plain
sentences are used in the learning algorithms and the resulting structure is
compared to the structure of the original corpus.

All ABL methods are tested ten times. The ABL:incr method is applied to
random orders of the input corpus. The probabilistic ABL methods select
constituents at random when different combinations of constituents have the
same probability. The results in table~\ref{fig:results} show the mean and
standard deviations (between brackets).

The two baseline systems, left and right, only build left and right branching
trees respectively.

Three metrics have been computed. \emph{NCBP} stands for Non-Crossing Brackets
Precision, which denotes the percentage of \emph{learned} constituents that do
not overlap with any constituents in the \emph{original} corpus. \emph{NCBR}
is the Non-Crossing Brackets Recall and shows the percentage of constituents in
the \emph{original} corpus that do not overlap with any constituents in the
\emph{learned} corpus. Finally, \emph{ZCS} stands for Zero-Crossing Sentences
and represents the percentage of \emph{sentences} that do not have any
overlapping constituents.

\subsection{Evaluation}

The \emph{incr} model performs quite well considering the fact that it
cannot recover from incorrect constituents, with a precision and recall of over
80~\%. The order of the sentences however is quite important, since the
standard deviation of the \emph{incr} model is quite high (especially with
the ZCS, reaching 3.22~\% on the OVIS corpus).

We expected the probabilistic methods to perform better, but the \emph{leaf}
model performs slightly worse. The ZCS, however, is somewhat better, resulting
in 21.63~\% on the ATIS corpus. Furthermore, the standard deviations of the
\emph{leaf} model (and of the \emph{branch} model) are close to 0~\%. The
statistical methods generate more precise results.

The \emph{branch} model clearly outperform all other models. Using more
specific statistics generate better results.

Although the results of the ATIS corpus and OVIS corpus differ, the conclusions
that can be reached are similar.

\subsection{ABL Compared to Other Methods}

It is difficult to compare the results of the ABL model against other methods,
since often different corpora or metrics are used. The methods described by
\newcite{bib:iorfpbc} comes reasonably close to ours. The \emph{unsupervised}
method learns structure on plain sentences from the ATIS corpus resulting in
37.35~\% precision, while the \emph{unsupervised} ABL significantly
outperforms this method, reaching 85.31~\% precision. Only their
\emph{supervised} version results in a slightly higher precision of 90.36~\%.

The system that simply builds right branching structures results in 82.70~\%
precision and 92.91~\% recall on the ATIS corpus, where ABL got 85.31~\% and
89.31~\%. This was expected, since English is a right branching language; a
left branching system performed much worse (32.60~\% precision and 76.82~\%
recall). Conversely, right branching would not do very well on a Japanese
corpus (a left branching language). Since ABL does not have a preference for
direction built in, we expect ABL to perform similarly on a Japanese corpus.

\section{Discussion and Future Extensions}

\subsection{Recursion}
\label{s:recursion}

\begin{figure*}
\begin{tabular}{ll}
\textbf{learned} &
\textit{Please explain the (field FLT DAY in the (table)$_{13}$)$_{13}$}\\
\textbf{original} &
\textit{Please explain (the field FLT DAY in (the table)$_{NP}$)$_{NP}$}\\
\textbf{learned} &
\textit{Explain classes QW and (QX and (Y)$_{52}$)$_{52}$}\\
\textbf{original} &
\textit{Explain classes ((QW)$_{NP}$ and (QX)$_{NP}$ and (Y)$_{NP}$)$_{NP}$}\\
\end{tabular}
\caption{Recursive structures learned in the ATIS corpus}
\label{fig:recursive}
\end{figure*}

All ABL methods described here can learn recursive structures and have been
found when applying ABL to the ATIS and OVIS corpus. As can be seen in
figure~\ref{fig:recursive}, the learned recursive structure is similar to the
original structure. Some structure has been removed to make it easier to see
where the recursion occurs.

Roughly, recursive structures are built in two steps. First, the algorithm
generates the structure with different non-terminals. Then, the two
non-terminals are merged as described in section~\ref{s:ec}. The merging of
the non-terminals may occur anywhere in the corpus, since \emph{all} merged
non-terminals are updated.

\subsection{Wrong Syntactic Type}
\label{s:wst}

In section~\ref{s:ec} we made the assumption that a constituent in a certain
context can only have one type. This assumption introduces some problems.

The sentence \textit{John likes visiting relatives} illustrates such a problem.
The constituent \textit{visiting relatives} can be a noun phrase or a verb
phrase.

Another problem is illustrated in figure~\ref{fig:wellmeat}. When applying the
ABL learning algorithm to these sentences, it will determine that
\textit{morning} and \textit{nonstop} are of the same type. Unfortunately,
\textit{morning} is a noun, while \textit{nonstop} is an
adverb.\footnote{Harris's implication does hold in these sentences.
\textit{nonstop} can also be replaced by for example \textit{cheap} (another
adverb) and \textit{morning} can be replaced by \textit{evening} (another
noun).}

\begin{figure}
\textit{\underline{Show me the} ( morning )$_X$ \underline{flights}}\\
\textit{\underline{Show me the} ( nonstop )$_X$ \underline{flights}}
\caption{Wrong syntactic type}
\vspace{-.2cm}
\label{fig:wellmeat}
\end{figure}

A future extension will not only look at the type of the constituents, but also
at the context of the constituents. In the example, the constituent
\textit{morning} may also take the place of a subject position in other
sentences, but the constituent \textit{nonstop} never will. This information can
be used to determine when to merge constituent types, effectively loosening the
assumption.

\subsection{Weakening Exact Match}

When the ABL algorithms try to learn with two completely distinct sentences,
nothing can be learned. If we weaken the exact match between words in the
alignment step of the algorithm, it is possible to learn structure even with
distinct sentences.

Instead of linking exactly matching words, the algorithm should match words
that are equivalent. An obvious way of implementing this is by making use of
\emph{equivalence classes}. (See for example \cite{bib:diapcfasc}.) The idea
behind equivalence classes is that words which are closely related are grouped
together.

A big advantage of equivalence classes is that they can be learned in an
unsupervised way, so the resulting algorithm remains unsupervised.

Words that are in the same equivalence class are said to be sufficiently
equivalent, so the alignment algorithm may assume they are similar and may thus
link them. Now sentences that do not have words in common, but do have words
in the same equivalence class in common, can be used to learn structure.

When using equivalence classes, more constituents are learned and more
terminals in constituents may be seen as similar (according to the equivalence
classes). This results in a much richer structured corpus.

\subsection{Alternative Statistics}

At the moment we have tested two different ways of computing the probability of
a constituent: \emph{ABL:leaf} which computes the probability of the occurrence
of the terminals in a constituent, and \emph{ABL:branch} which computes the
probability of the occurrence of the terminals together with the root
non-terminal in a constituent, based on the learned corpus.

Of course, other models can be implemented. One interesting possibility takes a
DOP-like approach \cite{bib:bgaebtol}, which also takes into account the inner
structure of the constituents.

\section{Conclusion}

We have introduced a new grammar learning algorithm based on comparing
and aligning plain sentences; neither pre-labelled or bracketed sentences, nor
pre-tagged sentences are used. It uses distinctions between sentences to
find possible constituents and afterwards selects the most probable ones. The
output of the algorithm is a structured version of the corpus.

By taking entire sentences into account, the context used by the model is not
limited by window size, instead arbitrarily large contexts are used.
Furthermore, the model has the ability to learn recursion.

Three different instances of the algorithm have been applied to two corpora of
different size, the ATIS corpus (716 sentences) and the OVIS corpus (6,797
sentences), generating promising results. Although the OVIS corpus is almost
ten times the size of the ATIS corpus, these corpora describe a small
conceptual domain. We plan to apply the algorithms to larger domain corpora in
the near future.

\bibliographystyle{acl}
\bibliography{paper}

\end{document}